\documentclass[11pt,a4paper]{article}
\usepackage[hyperref]{acl2021}
\usepackage{times}
\usepackage{latexsym}

\usepackage{natbib}
\usepackage{graphicx}
\usepackage{subfigure}
\usepackage{amsmath}
\usepackage{amsfonts}
\usepackage{amssymb}
\usepackage{booktabs}
\usepackage{enumerate}
\usepackage{microtype}

\aclfinalcopy 
\def\aclpaperid{2283} 

\title{De-biasing Distantly Supervised Named Entity Recognition \\via Causal Intervention}

\author{
  Wenkai Zhang\textsuperscript{\rm 1,3},
  Hongyu Lin\textsuperscript{\rm 1,}\thanks{~ Corresponding authors},
  Xianpei Han\textsuperscript{\rm 1,2,}\footnotemark[1],
  Le Sun \textsuperscript{\rm 1,2} \\
  \textsuperscript{\rm 1}Chinese Information Processing Laboratory ~ 
  \textsuperscript{\rm 2}State Key Laboratory of Computer Science \\
  Institute of Software, Chinese Academy of Sciences, Beijing, China\\
  \textsuperscript{\rm 3} University of Chinese Academy of Sciences, Beijing, China \\
  \texttt{\{wenkai2019,hongyu,xianpei,sunle\}@iscas.ac.cn}
}

\date{}

\begin{document}
\maketitle
\begin{abstract}
  Distant supervision tackles the data bottleneck in NER by automatically generating training instances via dictionary matching. 
  Unfortunately, the learning of DS-NER is severely dictionary-biased, which suffers from spurious correlations and therefore undermines the effectiveness and the robustness of the learned models. In this paper, we fundamentally explain the dictionary bias via a Structural Causal Model (SCM), categorize the bias into intra-dictionary and inter-dictionary biases, and identify their causes. 
  Based on the SCM, we learn de-biased DS-NER via causal interventions. For intra-dictionary bias, we conduct backdoor adjustment to remove the spurious correlations introduced by the dictionary confounder. For inter-dictionary bias, we propose a causal invariance regularizer which will make DS-NER models more robust to the perturbation of dictionaries. Experiments on four datasets and three DS-NER models show that our method can significantly improve the performance of DS-NER.
\end{abstract}

\section{Introduction}
Named entity recognition (NER) aims to identify text spans pertaining to specific semantic types, which is a fundamental task of information extraction, and enables various downstream applications such as Relation Extraction \citep{lin2016re} and Question Answering \citep{bordes2015qa}.
The past several years have witnessed the remarkable success of supervised NER methods using neural networks \citep{lample2016blstmcrf, ma-hovy-2016-end,lin-etal-2020-rigorous}, which can automatically extract effective features from data and conduct NER in an end-to-end manner.
Unfortunately, supervised methods rely on high-quality labeled data, which is very labor-intensive, and thus severely restricts the application of current NER models.
To resolve the data bottleneck, a promising approach is distant supervision based NER (DS-NER).
DS-NER automatically generates training data by matching entities in easily-obtained dictionaries with plain texts.
Then this distantly-labeled data is used to train NER models, commonly be accompanied by a denoising step.
DS-NER significantly reduces the annotation cost for building an effective NER model, and therefore has attracted great attention in recent years \citep{yang2018rl, shang2018tob, peng2019pul, cao2019lrnt, liang2020bond, Zhang_Lin_Han_Sun_Liu_Wei_Yuan_2021}.

\begin{figure}
  \centering
  \includegraphics[width=0.98\linewidth]{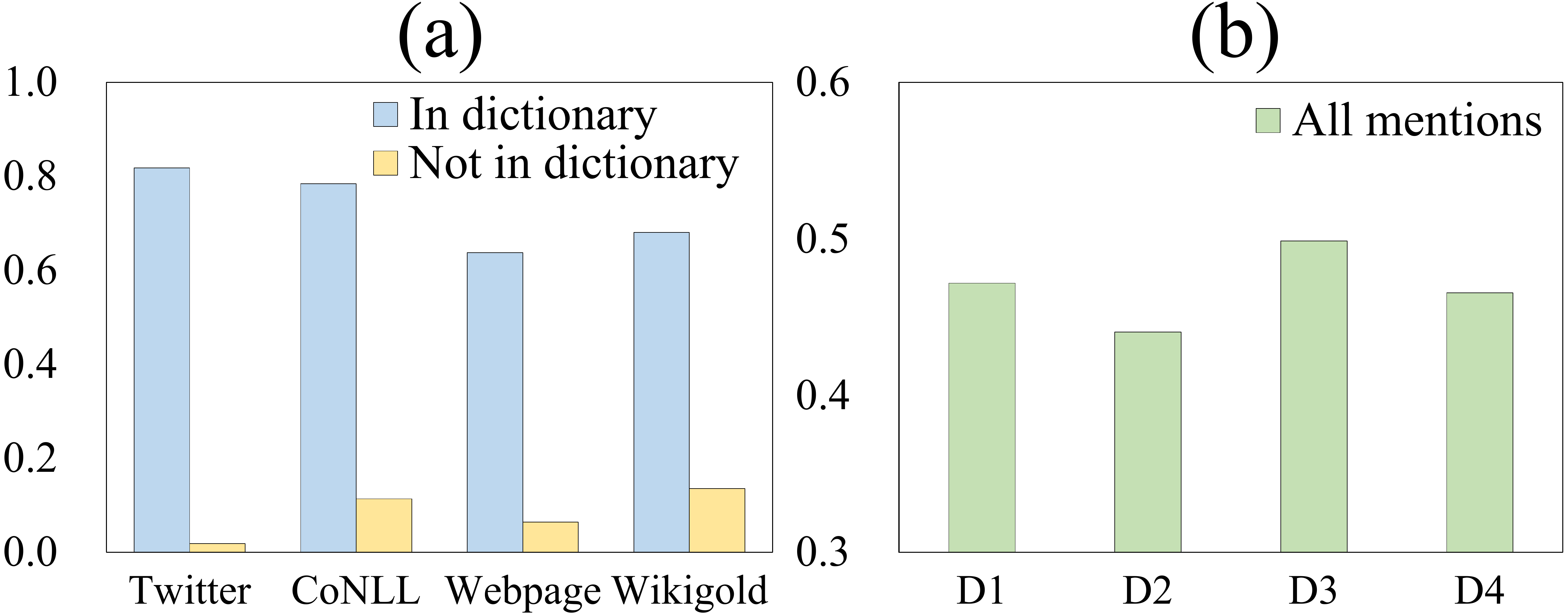}
  \caption{Dictionary bias in DS-NER happens both at intra-dictionary and inter-dictionary aspects: (a) Averaged likehoods of mentions in/not in the dictionary significantly diverge. (b) Mention likehoods of the models using different dictionaries significantly diverge.}
  \label{intra-inter-bias}
\end{figure}

\begin{figure*}[ht]
  \centering
  \includegraphics[width=0.8\linewidth]{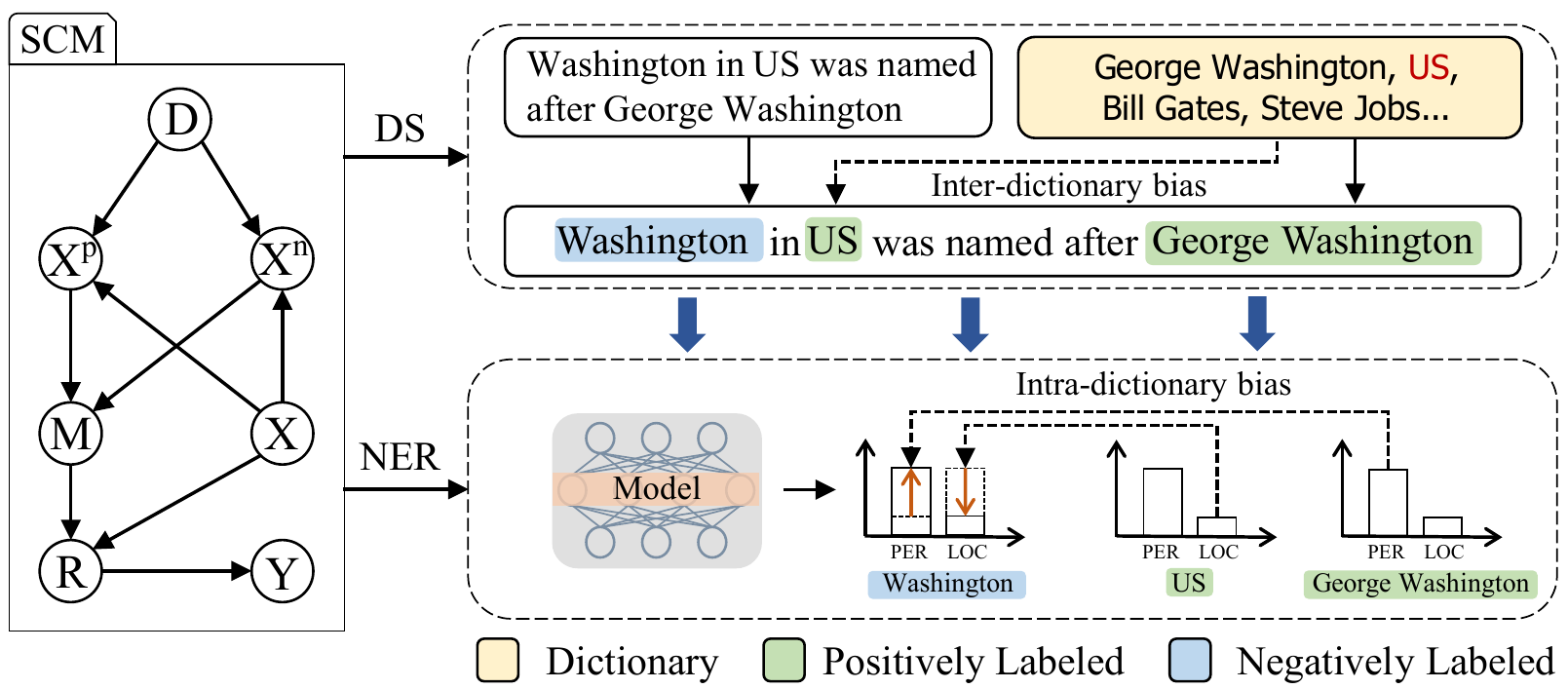}
  \caption{The proposed structural causal model for DS-NER. It can be roughly divided into two parts: distant supervision (DS) and NER. From the SCM, we identify that the intra-dictionary bias stems from the spurious correlations caused by backdoor paths, while the inter-dictionary bias stems from the over-fit on the dictionary characteristics. Detailed explanations can be found in Section~\ref{scm_dsner}.}
  \label{causal_graph}
\end{figure*}

However, the learning of DS-NER is \emph{dictionary-biased}, which severely harms the generalization and the robustness of the learned DS-NER models.
Specifically, entity dictionaries are often incomplete (missing entities), noisy (containing wrong entities), and ambiguous (a name can be of different entity types, such as \emph{Washington}). And DS will generate positively-labeled instances from the in-dictionary names but ignore all other names.
Such a biased dataset will inevitably mislead the learned models to overfit in-dictionary names and underfit out-of-dictionary names. We refer to this as \emph{intra-dictionary bias}.
To illustrate this bias, Figure~\ref{intra-inter-bias} (a) shows the predicting likelihood of a representative DS-NER model (RoBERTa + Classifier \citep{liang2020bond}).
We can see that there is a remarkable likelihood gap between in-dictionary mentions and out-of-dictionary mentions: the average likelihoods of out-of-dictionary mentions are $<0.2$, which means that a great majority of them cannot be recalled.
Furthermore, such a skewed distribution makes DS-NER models very sensitive to slight perturbations.
We refer to this as \emph{inter-dictionary bias}, i.e., different dictionaries can result in very different model behaviors.
In the example shown in Figure~\ref{intra-inter-bias} (b), we train the same DS-NER model by respectively using 4 dictionaries sampled from the same original dictionary, where each of them covers 90\% of entities in the original one.
We can see that the predicting likelihood diverges significantly even these 4 dictionaries share the majority part. 
Consequently, the dictionary-biased learning will undermine both the effectiveness and robustness of DS-NER models.

In this paper, we propose a causal framework to fundamentally explain and resolve the dictionary bias problem in DS-NER.
We first formulate the procedure of DS-NER from the causal view with a Structural Causal Model (SCM)~\cite{pearl2000models}, which is shown in the left part of Figure~\ref{causal_graph}.
From the SCM, we identified that the intra-dictionary bias stemming from the dictionary which serves as a confounder during the model learning.
The dictionary confounder will introduce two backdoor paths, one from positively-labeled instances ($X^p$) to entity labels ($Y$) and the other from negatively-labeled instances ($X^n$) to entity labels.
These backdoor paths introduce spurious correlations during learning, therefore result in the intra-dictionary bias.
Furthermore, the current learning criteria of DS-NER models is to optimize over the correlations between the instances ($X$) and entity types ($Y$) given one specific dictionary ($D$), namely $P(Y|X,D)$. Such criteria, however, diverges from the primary goal of learning a dictionary-free NER model (i.e., $P(Y|X)$), and results in the inter-dictionary bias.
Based on the above analysis, unbiased DS-NER should remove the spurious correlations introduced by backdoor paths and capture the true dictionary-free causal relations.

To this end, we conduct causal interventions to de-bias DS-NER from the biased dictionary.
For intra-dictionary bias, we intervene on the positive instances and the negative instances to block the backdoor paths in SCM, then the spurious correlations introduced by dictionary confounder will be removed.
Specifically, we conduct backdoor adjustment to learn de-biased DS-NER models, i.e., we optimize the DS-NER model based on the causal distribution, rather than from the spurious correlation distribution. 
For inter-dictionary bias, we propose to leverage causal invariance regularizer~\cite{mitrovic2021representation}, which will make the learned representation more robust to the perturbation of dictionaries.
For each instance in the training data, causal invariance regularizer will preserve the underlying causal effects unchanged across different dictionaries.
The proposed method is model-free, which can be used to resolve the dictionary bias in different DS-NER models by being applied as a plug-in during model training.

We conducted experiments on four standard DS-NER datasets: CoNLL2003, Twitter2005, Webpage, and Wikigold.
Experiments on three state-of-the-art DS-NER models show that the proposed de-biasing method can effectively solve both intra-dictionary and inter-dictionary biases, and therefore significantly improve the performance and the robustness of DS-NER in almost all settings.
Generally, the main contributions of this paper are:
\begin{itemize}
  \item We proposed a causal framework, which not only fundamentally formulates the DS-NER process, but also explains the causes of both intra-dictionary bias and inter-dictionary bias.
  \item Based on the causal framework, we conducted causal interventions to de-bias DS-NER. For intra-dictionary bias, we conduct causal interventions via backdoor adjustment to remove spurious correlations introduced by the dictionary confounder. For inter-dictionary bias, we propose a causal invariance regularizer which will make DS-NER models more robust to the perturbation of dictionaries.
  \item Experimental results on four standard DS-NER datasets and three DS-NER models demonstrate that our method can significantly improve the performance and the robustness of DS-NER.
\end{itemize}

\section{A Causal View on DS-NER}
\label{scm_dsner}

In this section, we formulate DS-NER with a structural causal model (SCM), then identify the causes of both intra-dictionary bias and inter-dictionary bias using the SCM.
An SCM captures the causal effect between different variables and describes the generative process of a causal distribution, which can be visually presented using a directed acyclic graph (DAG).
In SCM, each node represents a random variable, and a directed edge represents a direct causal relationship between two variables.
Based on SCM, the confounders and backdoor paths~\cite{pearl2000models} can be identified.
In the following, we will describe the causal view of DS-NER and then identify the dictionary bias.

\subsection{Structural Causal Model for DS-NER}
Figure \ref{causal_graph} shows the structural causal model for DS-NER, which contains 7 key variables in the DS-NER procedure:
1) \textbf{the applied dictionary $D$} for distant annotation;
2) \textbf{the unlabeled instances $X$}, where each instance is a pair of (mention candidate, context), and in training stage $X$ will be automatically labeled by $D$;
3) \textbf{the positive training instances $X^p$}, which are instances in $X$ being labeled as positive instances (i.e., entity mentions) by dictionary $D$;
4) \textbf{the negative training instances $X^n$}, which are instances being labeled as negative instances by dictionary $D$;
5) \textbf{the learned DS-NER model $M$}, which summarizes NER evidences from DS-labeled data during training, and predicts new instances during testing;
6) \textbf{the representations of instances $R$}, which is encoded dense representations of instances $X$ using the learned model $M$;
7) \textbf{the predicted entity labels $Y$} of instances in $X$ based on the representation $R$.

Defining these variables, the causal process of DS-NER can be formulated using SCM into two steps: distant supervision (DS) step and NER step respectively. For DS step, the procedure will generate DS-labeled data and learn DS-NER models by following causal relations:
\begin{itemize}
  \item ${D} {\rightarrow}{X^p} {\leftarrow} {X}$ and ${D} {\rightarrow}{X^n} {\leftarrow} {X}$ represent the distant annotation process, which uses dictionary $D$ to annotate the unlabeled instances $X$ and splits them into two sets: $X^p$ and $X^n$.
  \item ${X^p} {\rightarrow}{M} {\leftarrow} {X^n}$ represents the learning process, where model $M$ is the learned DS-NER model using $X^p$ and $X^n$. We denote the $X^p$ and $X^n$ generated from dictionary $D$ as $X^p(D)$ and $X^n(D)$ respectively.
\end{itemize}
And the causal relation in NER step can be summarized as:
\begin{itemize}
  \item ${M} {\rightarrow}{R} {\leftarrow} {X}$ is the representation learning procedure, which uses the learned model $M$ to encode instances $X$.
  \item ${R} {\rightarrow} {Y}$ represents the entity recognition process, where the labels of instances depend on the learned representation $R$ and instances $X$. We denote the entity labels corresponding to $X^p$ and $X^n$ as $Y^p$ and $Y^n$ respectively.
\end{itemize}

\subsection{Cause of Intra-dictionary Bias}
\label{intra_dict_bias}
Given distant annotation $X^p$ and $X^n$, the learning process of DS-NER will maximize the probability $P(Y^p{=}1, Y^n{=}0|X^p,X^n,D)$.
Unfortunately, because $D$ is a confounder for $X^p$ and $X^n$ in SCM, this criteria will introduce spurious correlations and result in the intra-dictionary bias:
(1) When maximizing $P(Y {=} 1|X^p,D)$, we want NER models to rely on the actual causal path ${X^p}{\rightarrow}{Y}$.
However, in SCM there exists a backdoor path ${X^p}{\leftarrow}{D}{\rightarrow}{X^n}{\rightarrow}{M}$ which will introduce spurious correlation between $Y$ and $X^p$.
Intuitively, this backdoor path appears as the false negative instances in $X^n$.
Because these false negative instances have correct entity contexts but out-of-dictionary names, they will mislead the models to underfit the entity context for prediction.
(2) When maximizing $P(Y {=} 0|X^n,D)$, we want NER models to rely on the actual causal path ${X^n}{\rightarrow}{Y}$.
However, in SCM there exists a backdoor path${X^n}{\leftarrow}{D}{\rightarrow}{X^p}{\rightarrow}{M}$ which will introduce spurious correlation between $Y$ and $X^n$.
Intuitively, this backdoor path appears as the false positive instances in $X^p$.
Because these false positive instances have in-dictionary entity names but spurious context, they will mislead the models to overfit the names in dictionary.

In general, the intra-dictionary bias is caused by backdoor paths introduced by $D$, and this bias will mislead the NER models to overfit names in dictionary and underfit the context of entities.

\subsection{Cause of Inter-dictionary Bias}

As mentioned above, DS-NER models are learned by fitting $P(Y^p{=}1,Y^n{=}0|X^p,X^n,D)$.
This criteria will mislead the model when learning the correlation between $X$ and $Y$ with spurious information in $D$ because the learning criteria is conditioned on it. 
However, a robust NER model should fit the underlying distribution $P(Y | X)$, rather than the dictionary-conditioned distribution $P(Y|X,D)$.
From the SCM, the dictionary $D$ will significantly influence the learned NER models $M$, and in turn result in different learned causal effects in the path $X\rightarrow R\rightarrow Y$ and entity prediction $Y$.
As a result, DS-NER models will fit different underlying distributions given different dictionaries, and therefore results in inter-dictionary bias.

However, in real-world applications, the dictionaries are affected by various factors, such as source, coverage or time.
Therefore, to enhance the robustness of the learning process, it is critical to alleviate the spurious influence of dictionary $D$ on the learned causal effects between $X$ and $Y$.
That is, we want DS-NER models to capture the dictionary-invariant entity evidence, rather than fit the dictionary-specific features.

\section{De-biasing DS-NER via Causal Intervention}

In this section, we describe how to de-bias DS-NER. Specifically, for intra-dictionary bias, we propose to use backdoor adjustment to block the backdoor paths. For inter-dictionary bias, we design a causal invariance regularizer to capture the dictionary-invariant evidence for NER.

\subsection{Removing Intra-dictionary Bias via Backdoor Adjustment}
\label{ramove_intra_bias}

Based on the analysis in Section \ref{intra_dict_bias}, the intra-dictionary bias is caused by the backdoor paths ${X^p}{\leftarrow}{D}{\rightarrow}{X^n}{\rightarrow}{M}$ and ${X^n}{\leftarrow}{D}{\rightarrow}{X^p}{\rightarrow}{M}$.
To remove these biases, we block both backdoor paths by intervening both $X^p$ and $X^n$.
After causal intervention, the learning of DS-NER models will fit the correct causal relation $P(Y^p{=}1|do( X^p(D) ),X^n)$ and  $P(Y^n{=}0|do(X^n(D)),X^p)$. Here $do(X^p(D)){=}do(X^p{=}X^p(D))$ represents the mathematical operation to intervene $X^p$ and preserve it to be $X^p(D)$ in the whole population.

\paragraph{Backdoor Adjustments.} To calculate the distribution $P(Y^p{=}1|do(X^p(D)))$ after causal intervention, we conduct backdoor adjustment according to causal theory~\cite{pearl2009causal}:
\begin{equation}
  \begin{aligned}
    P_{pos}(D) {\triangleq} &P(Y^p{=}1|do(X^p(D))) \\
               {=} &\sum_{i} P(Y^p{=}1|X^p(D), X^n(D_i)) \\
               &\quad\ \ \times P(D_i)
    \label{e1}
  \end{aligned}
\end{equation} 
where $X^n(D_i)$ denotes the negative instances generated from the DS dictionary $D_i$. $P(Y^p{=}1|X^p(D), X^n(D_i))$ is the probability of predicting $X^p(D)$ into $Y{=}1$, which can be formulated using a neural network-based DS-NER model parametrized by $\theta$, i.e., $P(Y|X^p, X^n)$  =  $P(Y |X^p, X^n;\theta)$. Detailed derivations is shown in appendix \ref{appendix:proof}.

\begin{figure}
  \centering
  \includegraphics[width=0.98\linewidth]{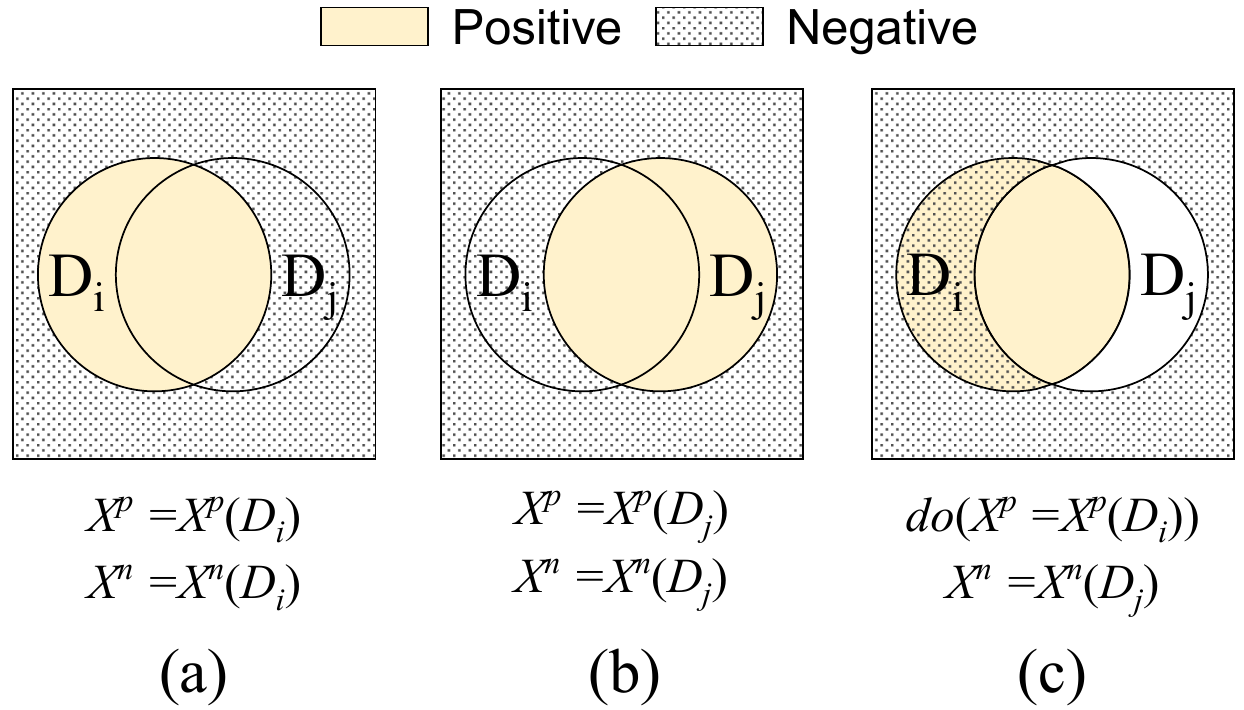}
  \caption{An illustration on causal intervention. (a) and (b) show the generated instances according to dictionary $D_i$ and $D_j$. And (c) shows the generated instances from $do(X^p(D_i))$ and $X^{n}(D_j)$.}
  \label{ba}
\end{figure}

Note the distribution $P(Y^p{=}1 | do(X^p(D)))$ in the causal framework is not the marginalized distribution $P(Y^p{=}1 | X^p(D))$ in the probability framework. Otherwise the marginalization should take place in the conditional distribution $P(D_i|X^p)$ rather than $P(D_i)$. Furthermore, as shown in Figure~\ref{ba}, $X^p {=} X^p(D_i)$ and $X^n {=} X^n(D_j)$ can not happen together in probabilistic view unless $D_i {=} D_j$. However, in the causal view, they can happened together via the causal intervention. That is $do(X^p {=} X^p(D_i))$ and $X^n {=} X^n(D_j)$, which is shown in Figure~\ref{ba} (c). For more details, please refer to~\cite{neal2020cau} for a brief introduction.

Similarly, to block the backdoor paths and calculate the causal distribution $P(Y^n{=}0|do(X^n(D)))$, we can conduct backdoor adjustment on $X^n$ by:
\begin{equation}
  \begin{aligned}
    P_{neg}(D) {\triangleq} & P(Y^n{=}0|do(X^n(D))) \\
               {=}& \sum_{i} P(Y^n{=}0 |X^n(D), X^p(D_i)) \\
                &\quad\ \ \times P(D_i)
    \label{e2}
  \end{aligned}
\end{equation}
\paragraph{Estimating Dictionary Probabilities. } Because we only have one global dictionary $D$, it is hard to estimate the probability of other dictionary $D_i$ used in the Equation~(\ref{e1}) and~(\ref{e2}). To tackle this problem, we sample $K$ sub-dictionaries by sampling entities from the global dictionary $D$. The probability of each entity being sampled corresponds to its utterance frequency in a large-scale corpus.
Then we applied a uniform probability assumption to these sampled dictionaries, which means that these sub-dictionaries will then be used to conduct backdoor adjustment with equal dictionary probabilities, i.e., $P(D_i)=\frac{1}{K}$.

\paragraph{Learning DS-NER Models with Causal Relation. } Given the above two causal distributions after backdoor adjustment, the DS-NER models can be effectively learned, and the intra-dictionary bias can be eliminated based on the causal relations between $X^p$, $X^n$ and $Y$. Formally, we optimize DS-NER models by minimizing the following negative likelihood based on causal relation:
\begin{equation}
  L_{BA}(\theta) {=} - \log P_{pos}(D) - \log P_{neg}(D)
  \label{l1}
\end{equation}

Note that the proposed method is model-free, which means that it can be applied to the majority of previous DS-NER methods by adaptively changing the underlying parametrization of probability distribution $P(Y|X^p, X^n;\theta)$ .

\subsection{Eliminating Inter-dictionary Bias via Causal Invariance Regularizer}
This section describes causal invariance regularizer to eliminate the inter-dictionary bias. Specifically, after backdoor adjustment for intra-dictionary bias, the causal distribution we optimize (i.e., $P_{pos}(D)$ and $P_{neg}(D)$) still depends on the dictionary $D$. As a result, given different dictionaries, DS-NER models will fit different underlying causal distributions and result in inter-dictionary bias.

Ideally, a robust DS-NER learning algorithm should be dictionary-free, i.e., we should directly optimize towards the implicit distribution of $P(Y|X)$. However, it is impossible to directly achieve this because the golden answer $Y$ of $X$ is invisible in DS-NER. To enhance the robustness of the learning process, this section proposes a causal invariance regularizer, which ensures DS-NER models to learn useful entity evidence for NER but not to fit dictionary-specific features. Specifically, the goal of causal invariance~\cite{pearl2000models} is to ensure learned NER models will keep similar causal effects using different dictionaries, which can be formulated as:
\begin{equation}
  \begin{aligned}
    \theta^{*}_{inv} {=} \arg\min_{\theta} \| &P_{pos}(D_i)-P_{pos}(D_j) \\
                                             +&P_{neg}(D_i)-P_{neg}(D_j) \|
  \end{aligned}
\end{equation}

Here $||*||$ measures the distance between two distributions. 
However, as we mentioned above, this distance cannot be directly optimized because the golden label $Y$ of $X$ is unknown. Fortunately, in the SCM, the impact from dictionary $D$ to the entity label $Y$ are all through the model $M$ and the representation $R$, i.e., through the path $D \rightarrow M \rightarrow R \rightarrow Y$. As a result, the bias from the dictionary $D$ can be eliminated by preserving the causal effects between $X$ and any node in the path. A simple and reasonable solution is to preserve the causal invariance of the representation $R$. That is, given different dictionaries, we keep the causal effects from $X$ to $R$ unchanged, and therefore causal effects of $X \rightarrow Y$ will remain unchanged. Specifically, when learning causal effects given an dictionary $D$, the causal invariance regularizer will further enhance its causal consistency with other dictionaries by minimizing its representation distances to other dictionaries:
\begin{align}
  L_{CIR}(\theta;D) {=} \sum_{i{=}1}^{K} \sum_{x \in X}
   & || R_{D}(x;\theta) \notag \\
   & -R_{D^i}(x) ||^2
  \label{l2}
\end{align}
Here $R_{D}(x;\theta)$ is the representation of instance $x$, which is derived from the NER model $M$ by fitting the causal effects of dictionary $D$. The reference dictionary $D^i$ in the formulations are generated in the same way as we described in Section~\ref{ramove_intra_bias} and $K$ is the number of sub-dictionaries. Therefore, this regularizer ensures that the representations learned using different dictionaries will be consistent, and the inter-dictionary bias is eliminated.

Finally, we combine (\ref{l1}) and (\ref{l2}) to de-bias both intra-dictionary bias and inter-dictionary bias and obtain the final DS-NER models by optimizing:
\begin{equation}
  L = \sum_{i} L_{BA}^i + \lambda L_{CIR}
\end{equation}
where $\lambda$ is a hyper-parameter which controls the relative importance of these two losses and is tuned on the development set.

\section{Experiments}

\subsection{Experimental Settings}

\paragraph{Datasets.} We conduct experiments on four standard datasets:
(1) CoNLL2003 \citep{conll2003} is a well known open-domain NER dataset. It consists of 20744 sentences collected from 1393 English news articles and is annotated with four types: PER, ORG, LOC and MISC.
(2) Twitter \citep{twitter2005} is from the WNUT 2016 NER shared task. It consists of 7236 sentences with 10 entity types.
(3) Webpage \citep{webpage2009} contains 20 webpages, including personal, academic and computer-science conference homepages. It consists of 619 sentences with the four types the same as CoNLL2003.
(4) Wikigold \citep{wikigold2009} contains 149 articles from the May 22, 2008 dump of English Wikipedia. It consists of 1969 sentences with the same types of CoNLL2003.

\paragraph{Distant Annotation Settings.}
We use two distant annotation settings: String-Matching and KB-Matching \citep{liang2020bond}.
String-Matching labels dataset by directly matching names in dictionary with sentences.
KB-Matching is more complex, which uses a set of hand-crafted rules to match entities.
We find KB-Matching can generate better data than String-Matching, but String-Matching is a more general setting.
In our experiments, we report performance on both KB-Matching and String-Matching settings.

\paragraph{Implementation Detail.}
We implement BiLSTM-CRF with AllenNLP \citep{Gardner2017AllenNLP}, an open-source NLP research library, and the input vector is the 100-dimension GloVe Embeddings \citep{glove2014}.
For other baselines, we use the officially released implementation from the authors. 
We openly release our source code at github.com/zwkatgithub/DSCAU.

\begin{table*}[!t]
  \centering
  \scalebox{0.68}{
    \begin{tabular}{lcccc|cccc}
      \toprule
                              & \multicolumn{4}{c}{KB-Matching} & \multicolumn{4}{c}{String-Matching}                                                                                           \\ \midrule
      Methods                 & CoNLL2003                       & Twitter                             & Webpage      & Wikigold     & CoNLL2003    & Twitter      & Webpage      & Wikigold     \\ \midrule
                              & \multicolumn{8}{c}{Supervised Learning Baselines} \\ \midrule
      BiLSTM-CRF              & 85.98                           & 32.30                               & 51.59        & 57.01        & --           & --           & --           & --           \\  
      RoBERTa-base            & 91.12                           & 50.47                               & 74.07        & 84.02        & --           & --           & --           & --           \\ \midrule
                              & \multicolumn{8}{c}{Distant Supervision Baselines} \\ \midrule
      DictMatch               & 71.40                           & 35.83                               & 52.45        & 47.76        & 43.91        & 19.18        & 2.56         & 19.04        \\
      BiLSTM-CRF              & 64.62                           & 30.25                               & 13.90        & 37.46        & 60.52        & 31.67        & 24.67        & 27.97        \\ \midrule
      RoBERTa-base            & 76.04                           & 46.40                               & 54.07        & 52.83        & 73.94        & 46.02        & 57.14        & 37.94        \\
      \textbf{\quad+BA} (Ours)     & 76.43(+0.40)                    & 46.75(+0.35)                        & 59.28(+5.21) & 53.56(+0.73) & 75.43(+1.49) & 46.69(+0.67) & 58.71(+1.57) & 42.23(+4.59) \\
      \textbf{\quad+BA+CIR} (Ours) & 78.78(+2.74)                    & 47.12(+0.72)                        & 59.06(+4.99) & 55.60(+2.77) & 75.59(+1.65) & 47.27(+1.25) & 58.04(+0.90) & 44.19(+6.25) \\ \midrule
      BOND                    & 79.83                           & 47.72                               & 61.28        & 60.23        & 75.51        & 48.72        & 66.23        & 42.17        \\
      \textbf{\quad+BA} (Ours)       & 80.81(+0.98)                    & 48.45(+0.73)                        & 64.65(+3.37) & 60.81(+0.58) & 76.21(+0.70) & 49.12(+0.40) & 66.67(+0.44) & 42.53(+0.36) \\
      \textbf{\quad+BA+CIR} (Ours)   & 81.54(+1.71)                    & 49.01(+1.29)                        & 64.71(+3.43) & 61.48(+1.25) & 76.53(+1.02) & 48.82(+0.10) & 66.67(+0.44) & 45.55(+3.38) \\ \bottomrule
    \end{tabular}}
  \caption{F1 scores on CoNLL2003, Twitter, Webpage and Wikigold. BA and CIR denotes the proposed backdoor adjustment and causal invariance regularizer respectively. We can see that the proposed causal intervention approach achieves significant improvements on almost all settings.}
  \label{main_reuslt}
\end{table*}

\begin{table}[!t]
  \centering
  \scalebox{0.82}{
    \begin{tabular}{lcc}
      \toprule
                          & KB-Matching  & String-Matching \\ \midrule
      PU-Learning                 & 74.96        & 72.42           \\
      \textbf{\quad+BA} (Ours)    & 80.93(+5.97) & 76.17(+3.75)    \\
      \textbf{\quad+BA+CIR} (Ours) & 81.96(+7.00) & 76.62(+4.20)    \\ \bottomrule
    \end{tabular}
  }
  \caption{F1 scores on CoNLL2003 dataset based on PU-Learning~\cite{peng2019pul}. We don't report the results on other datasets, because PU-Learning needs high-quality dictionary which only provided by CoNLL2003.}
  \label{pul}
\end{table}

\subsection{Baselines}
The proposed de-biased training strategy is both model-free, and learning algorithm-free.
Therefore, we use the following base DS-NER baselines and compare the performance of using/not using our de-biased training strategy:
\paragraph{DictMatch}, which perform NER by directly matching text with names in a dictionary, so no learning is needed.
\paragraph{Fully-supervised baselines}, including:
(i) BiLSTM-CRF \citep{lample2016blstmcrf}, which uses Glove \citep{glove2014} for word embeddings;
(ii) RoBERTa-base \citep{roberta}, which encodes text using RoBERTa-base then predict token label via a multi-layer perceptron.
\paragraph{Naive Distant Supervision (Naive)}, which directly uses weakly labeled data to train a fully-supervised model.
It could be considered as the lower bound of DS-NER.
\paragraph{Positive-Unlabeled Learning (PU-Learning)} \citep{peng2019pul}, which formulates DS-NER as a positive-unlabeled learning problem.
It could obtain unbiased loss estimation of unlabeled data.
However, it assumes that there are no false positive instances which may be incorrect in many datasets.
\paragraph{BOND} \citep{liang2020bond}, which is a two-stage learning algorithm: In the first stage, it leverages pre-trained language model to improve the recall and precison of the NER model; In the second stage, it adopts a self-training approach to further improve the model performance.

\subsection{Main Results}
Table \ref{main_reuslt} and Table \ref{pul} show the overall performance (F1 scores) of different baselines and our methods.
For our method, we use BA to denote backdoor adjustment, and CIR to denote causal invariance regularizer.
We conduct our debiasing method on three base models: RoBERTa-base, PU-Learning and BOND, therefore we have 6 systems of our methods:
RoBERTa+BA, RoBERT+BA+CIR, PU-Learning+BA, PU-Learning+BA+CIR, BOND +BA, BOND+BA+CIR.

We can see that:
(1) DS-NER models are severely influenced by the dictionary bias. Without debiasing, the naive DS-NER baselines BiLSTM-CRF and RoBERTa-base can only achieve comparable performance with the simple DictMatch baselines.
And by taking the dictionary bias into consideration, PU-Learning, BOND with our method can significantly improve the performance of DS-NER.
Compared with DictMatch, they correspondingly achieve 4.99\%, 21.98\% F1 improvements on average. This verified that the dictionary bias is critical for DS-NER models.
(2) By debiasing DS-NER models via causal intervention, our method can achieve significant improvement.
Compared with their counterparts, our full methods RoBERTa+BA+CIR, BOND+BA+CIR correspondingly achieve 4.91\%, 3.18\% improvements averaged on four datasets in KB-Matching (5.75\%, 2.56\% improvements on String-Matching) and PU-Learning+BA+CIR achieves 9.34\% improvement on CoNLL2003 dataset in KB-Matching (5.80\% improvement in String-Matching).
This verified the effectiveness of using causal intervention for debiasing DS-NER.
(3) Our method can effectively resolve both intra-dictionary and inter-dictionary biases. Both of backdoor adjustment and causal invariance regularizer can improve the NER performance.
By conducting backdoor adjustment, our method can achieve a 3.27\% F1 improvement averaged on all base models and all datasets.
And further conducting causal invariance regularizer can future improve 4.63\% average F1.

\subsection{Effects on Robustness}

\begin{figure}[!t]
  \centering
  \includegraphics[width=0.98\linewidth]{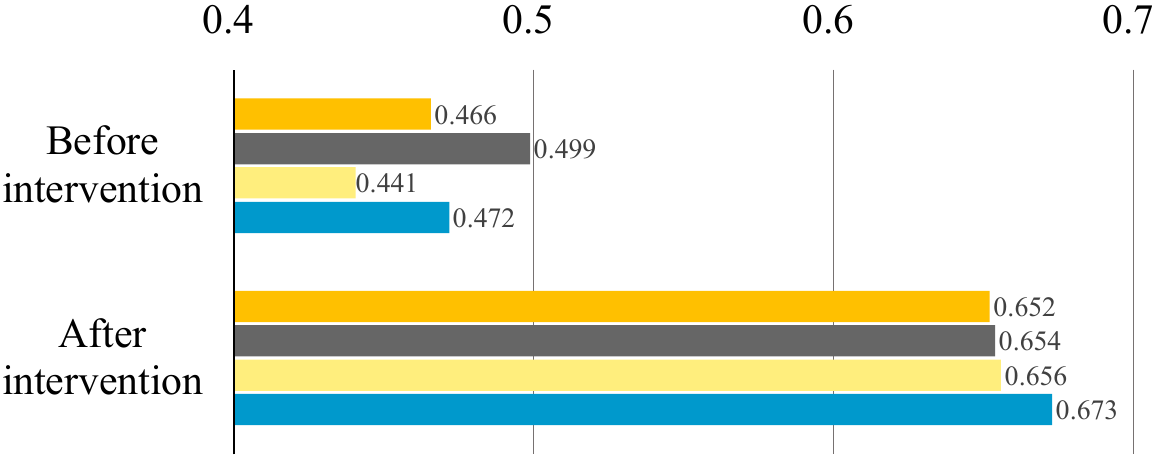}
  \caption{The likelihood variance between different dictionaries before/after using causal invariance regularizer (RoBERTa-Classifier on CoNLL2003), We can see that the performance variance significantly decreases, which verifies that causal invariance regularizer can significantly improve the robustness of DS-NER.}
  \label{veri_problem}
\end{figure}

To verify whether the causal invariance regularizer can significantly improve the robustness of DS-NER across different dictionaries, we further compared the predicting likelihood of golden mentions using different dictionaries. Specifically, we train the same RoBERTa-Classifier DS-NER models by sampling 4 dictionaries. Figure \ref{veri_problem} shows the average predicting likelihood before/after using our de-biasing method.

From Figure \ref{veri_problem}, we can see that the proposed causal invariance regularizer significantly reduced the likelihood gaps between different dictionaries. This verified that removing the inter-dictionary bias can significant benefit the robustness of DS-NER. Furthermore, we can see that the likelihoods of golden mentions are remarkably increased, which represents a better NER performance. These all demonstrate the effectiveness of the proposed causal invariance regularizer.

\begin{figure}[!t]
  \centering
  \includegraphics[width=0.75\linewidth]{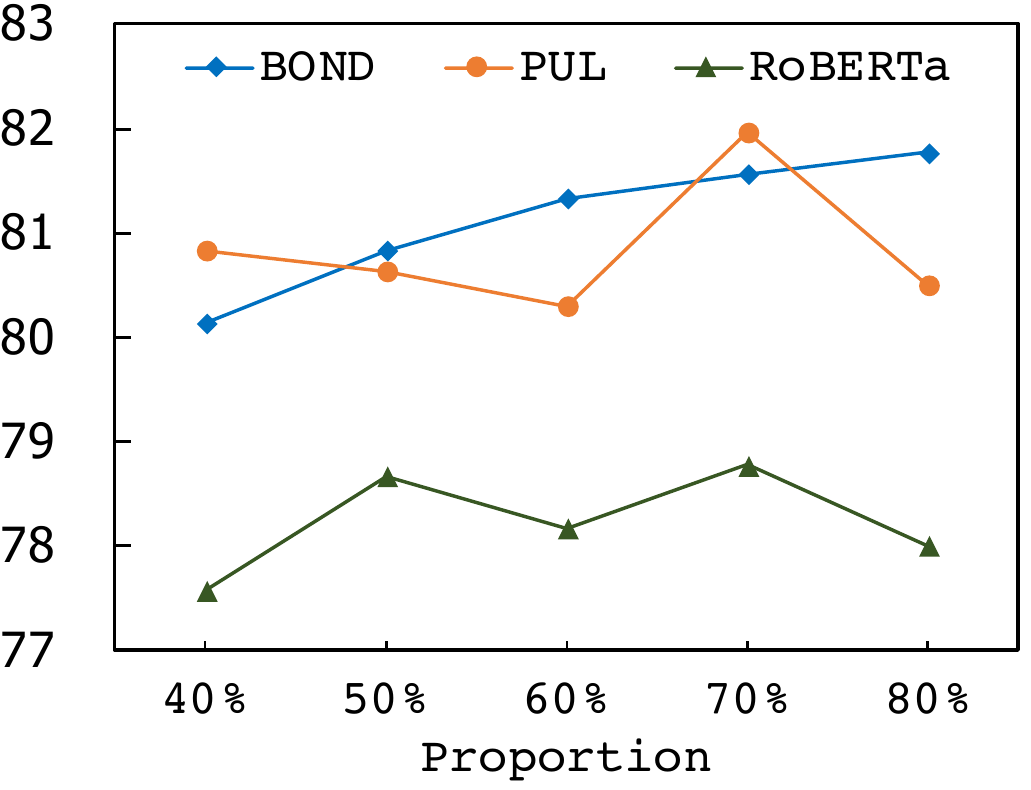}
  \caption{F1 scores when with different sub-dictionary coverages on the test set of CoNLL2003.}
  \label{prop}
\end{figure}
\begin{figure}[!t]
  \centering
  \includegraphics[width=0.75\linewidth]{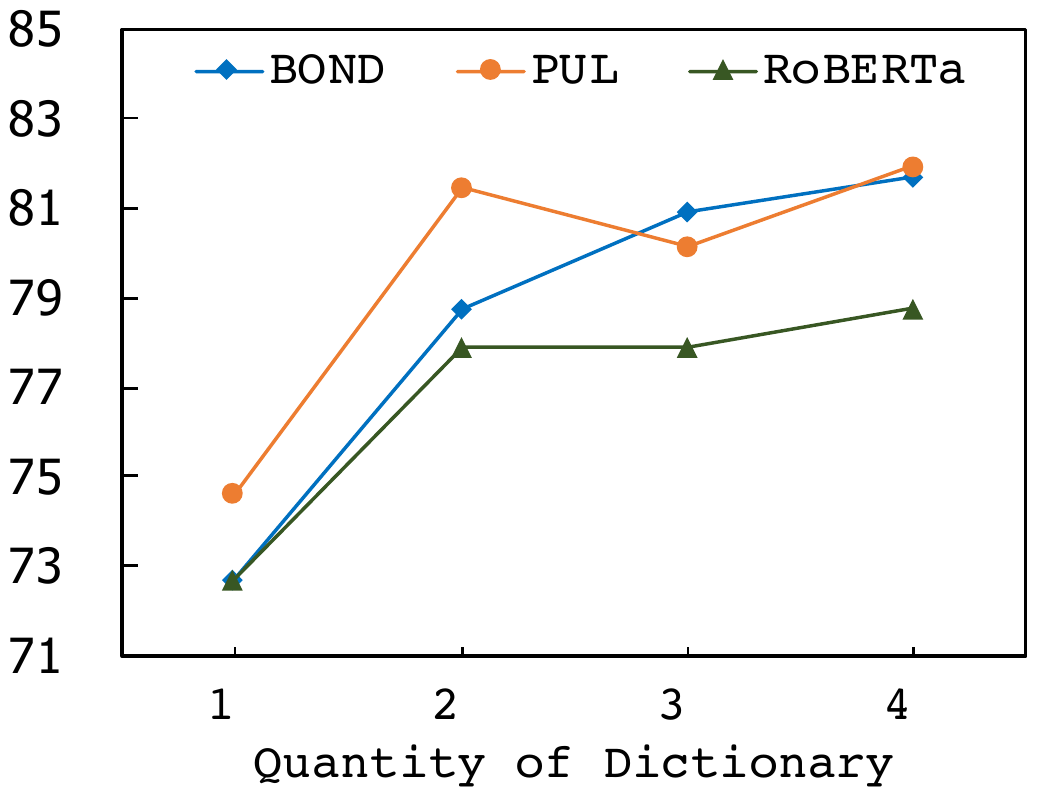}
  \caption{F1 scores when using different sub-dictionary quantities on the test set of CoNLL2003.}
  \label{dict_size}
\end{figure}

\subsection{Influence of Sub-dictionaries}
To conduct causal intervention, our method needs to sample sub-dictionaries from the original one.
To analyze the influence of the coverage and the quantity of sub-dictionaries, we conducted experiments on sub-dictionaries with different coverages and different quantities.

\paragraph{Dictionary Coverage.} Figure \ref{prop} shows the results with different dictionary coverages.
We can see that our method is not sensitive to the coverage of sub-dictionaries: it can achieve robust performance from 40\% to 80\% coverage.
All three models achieved the best performance at the 70\% coverage.
This result demonstrates the robustness of our method on dictionary coverage.

\paragraph{Dictionary Quantity.} Figure \ref{dict_size} shows the results with different sub-dictionary quantities.
We can see that our method can achieve performance improvement by sampling more sub-dictionaries.
This is because more sub-dictionaries will lead to more accurate estimation of both the dictionary probability in backdoor adjustment and the dictionary variance in causal invariance regularizer.
Futhermore, we can see that the performance using only one sub-dictionary (i.e., DS-NER without causal intervention) is significantly worse than other settings, this further verifies the effectiveness of our method.

\section{Related Work}
\noindent \textbf{DS-NER.} Supervised NER models have achieved promising performance \citep{lample2016blstmcrf, lin2019arn, lin-etal-2019-gazetteer}.
However, the reliance on labeled data limits their applications in open situations.
Distant supervision \citep{mintz2009ds} is a promising technique to alleviate the data bottleneck for NER, which generates large-scale training data by matching sentences with external dictionaries.
Current DS-NER studies focus on denoising the distantly labeled training data for better model learning.
\citet{yang2018rl} adopted reinforcement learning for denoising.
\citet{shang2018tob} proposed a sequence labeling framework TieOrBreak, which can avoid noise caused by a single word.
\citet{cao2019lrnt} promoted the quality of data by exploiting labels in Wikipedia.
\citet{peng2019pul} employed Positive-Unlabeled Learning to obtain unbiased estimation of the loss value.
\citet{liang2020bond} used self-training method which leverages a pretrained language model as teacher model to guide the training of student model.

\noindent \textbf{Causal Inference.} Causal Inference \citep{pearl2009causal, pearl2018why} has been widely adopted in psychology, politics and epidemiology for years \citep{mackinnon2007mediation, richiardi2013mediation, keele2015statistics}.
It can provide more reliable explanations by removing confounding bias in data, and also provide debiased solutions by learning causal effect rather than correlation effect.
Recently, many causal inference techniques are used in computer vision \citep{tang2020long, qi2020two} and natural language process \citep{wu2020biased, zeng2020counterfactual}.

\section{Conclusions}
This paper proposes to identify and resolve the dictionary bias in DS-NER via causal intervention. 
Specifically, we first formulate DS-NER using a structural causal model, then identity the causes of both intra-dictionary and inter-dictionary biases, finally de-bias DS-NER via backdoor adjustment and causal invariance regularizer.
Experiments on four datasets and three representative DS-NER models verified the effectiveness and the robustness of our method.

\section*{Acknowledgements}
This work is supported by the National Natural Science Foundation of China under Grants no.U1936207, Beijing Academy of Artificial Intelligence (BAAI2019QN0502), scientific research projects of the State Language Commission (YW135-78), and in part by the Youth Innovation Promotion Association CAS(2018141). Moreover, we thank all reviewers for their valuable comments and suggestions.

\bibliographystyle{acl_natbib}
\bibliography{acl2021}

\clearpage

\appendix

\section{Proof of Backdoor Adjustment}\label{appendix:proof}
We prove the backdoor adjustment for SCM using the do-calculus \citep{pearl1995causal} and the  Truncated Factorization \citep{neal2020cau}.

First of all, we write the joint distribution as shown in our causal graph:
\begin{equation}
  \begin{aligned}
    P(D,&X^p,X^n,Y,M,R,X) \notag \\
   = & P(D) P(X) P(X^p|D,X) P(X^n|D,X) \notag \\
  & P(M|X^p,X^n) P(R|M,X) P(Y|R)
  \end{aligned}
  \label{app1}
\end{equation}

Due to the objective of our method is debiasing DS-NER models during training, we ignore the unlabeled instances variable $X$ which is not related to the training process. Then we obtain the following equation:
\begin{equation}
  \begin{aligned}
    P(D,&X^p,X^n,Y,M,R) \notag \\
   = & P(D) P(X^p|D) P(X^n|D) \notag \\
  & P(M|X^p,X^n) P(R|M) P(Y|R)
  \end{aligned}
  \label{app2}
\end{equation}

Note that the prediction step of a NER model ${M}{\rightarrow}{R}{\rightarrow}{Y}$ doesn't have causal effects with other variables, we abbreviate $P(M|X^p,X^n) P(R|M,X) P(Y|R)$ as $P(Y|X^p, X^n)$.
Finally, we obtain the simplified joint distribution:
\begin{equation}
  \begin{aligned}
  P(&D, X^p,X^n, Y)  \notag \\
  = & P(D) P(X^p|D) P(X^n|D) P(Y|X^p,X^n)
  \end{aligned}
  \label{app3}
\end{equation}

Then we conduct causal intervention on $X^p$, i.e., $do(X^p{=}X^p(D))$ where $X^p(D)$ denotes positive instances generated by dictionary $D$. Here, we abbreviate it as $do(X^p(D))$. In practice, $do(X^p(D))$ denotes that we use these positive instances to calculate loss value, therefore, in order to explicitly indicate this, we use $Y^p{=}1$ in the following equation.
According to the Truncated Factorization \citep{neal2020cau}, we can know $P(X^p|D){=}1$, and obtain the following equation:
\begin{equation}
  \begin{aligned}
      P(D, &X^n, Y^p{=}1 | do(X^p(D)))  \notag \\
      = & P(D) P(X^n|D) P(Y^p{=}1|X^p(D),X^n)
  \end{aligned}
  \label{app4}
\end{equation}

Next, we integrate $D$ and $X^n$:
\begin{equation}
  \begin{aligned}
     & P(Y^p{=}1| do(X^p(D)))  \notag \\
    = &  \sum_i \sum_{X^n} P(D_i) P(X^n|D_i) P(Y^p{=}1|X^p(D),X^n) 
  \end{aligned}
  \label{app5}
\end{equation}
Note that $P(X^n(D_i)|D_i){=}1$ if and only if $X^n$ is generated by a specific dictionary $D_i$, therefore we can obtain:
\begin{equation}
  \begin{aligned}
     & P( Y^p{=}1 | do(X^p(D)))  \notag \\
      = & \sum_i \sum_{X^n} {P(D_i) P(X^n|D_i) P(Y^p{=}1|X^p(D),X^n)} \notag \\
      = & \sum_i { P(D_i) P(Y^p{=}1|X^p(D),X^n(D_i)) }
  \end{aligned}
  \label{app6}
\end{equation}

\end{document}